\documentclass[runningheads]{llncs}
\usepackage[T1]{fontenc}
\usepackage{amsfonts}
\usepackage{amsmath}

\usepackage{graphicx}
\usepackage[misc,geometry]{ifsym}
\usepackage{adjustbox}
\usepackage{amsmath, amssymb, bm}
\usepackage{hyperref}
\usepackage{color}

\usepackage[dvipsnames]{xcolor}
\usepackage{array, booktabs, multirow}
\newcolumntype{C}[1]{>{\centering\arraybackslash}c{#1}}
\usepackage{array}
\newcolumntype{C}[1]{>{\centering\arraybackslash}p{#1}}
\usepackage{algorithm}
\usepackage{algorithmic}
\usepackage{tabularx}
\usepackage{float}

\begin{document}
\title{Explainable Parallel RCNN with Novel Feature Representation for Time Series Forecasting}
\titlerunning{Explainable Parallel RCNN with Novel Feature Representation}
%
\author{Jimeng Shi \Letter \inst{1} \and
Rukmangadh Myana\inst{1} \and
Vitalii Stebliankin\inst{1} \and
Azam Shirali\inst{1} \and
Giri Narasimhan\inst{1}
}
\authorrunning{J. Shi et al.}
%
\institute{Knight Foundation School of Computing and Information Sciences, \\ Florida International University \\
\email{\{jshi008, rmyan001, vsteb002, ashir018, giri\}@fiu.edu}}
\maketitle              
\begin{abstract}
Accurate time series forecasting is a fundamental challenge in data science, as it is often affected by external covariates such as weather or human intervention, which in many applications, may be predicted with reasonable accuracy. We refer to them as \emph{predicted future covariates}. However, existing methods that attempt to predict time series in an iterative manner with auto-regressive models end up with exponential error accumulations. Other strategies that consider the past and future in the encoder and decoder respectively limit themselves by dealing with the past and future data separately. To address these limitations, a novel feature representation strategy - \emph{shifting} - is proposed to fuse the past data and future covariates such that their interactions can be considered. To extract complex dynamics in time series, we develop a parallel deep learning framework composed of RNN and CNN, both of which are used in a hierarchical fashion. We also utilize the \emph{skip connection} technique to improve the model's performance. Extensive experiments on three datasets reveal the effectiveness of our method. Finally, we demonstrate the model \emph{interpretability} using the Grad-CAM algorithm.

\end{abstract}
\section{Introduction}
\label{introduction}
Time series forecasting plays an essential role in many scenarios in real life. Accurate forecasting allows people to do better resource management \cite{niu2021evaluating} and optimization decisions \cite{cinar2017position} for critical processes. Applications include demand forecasting in retail \cite{bose2017probabilistic}, dynamic assignments of beds to patients \cite{zhang2018multi}, monthly inflation forecasting \cite{baybuza2018inflation}, and much more. Because of its popularity and significance, many time series forecasting methods have been explored. Traditional statistical forecasting methods, such as autoregression \cite{efendi2018}, exponential smoothing \cite{Hyndman2008Forecasting}, and ARIMA \cite{Chen2009ARIMA}, are widely utilized for univariate time series. These methods learn the temporal features (e.g., trends and seasonality) from past data and achieve good performance for univariate time series prediction. But they are ineffective to learn the complex dynamics among multivariate time series, partly because of their inability to take advantage of \emph{covariates} - independent variables that can influence the target variable, although perhaps not directly.

Good time series forecasting requires substantial amounts of historical data of the target variable(s) to learn temporal patterns.
They also require the exogenous covariates to learn the dependent relationships. 
More importantly, in many applications, some of the covariates can be predicted with reasonable accuracy for the immediate future. We refer to such covariates from the immediate future as \emph{predicted future covariates}. For example, in terms of the task predicting water levels in a river or canal system, a covariate of interest could be \emph{precipitation}. And it is possible to use historical data as well as reasonably accurate predictions for the near future, which may be obtained from the weather services. 
\begin{figure}
\begin{center}
\includegraphics[width=0.7\textwidth]{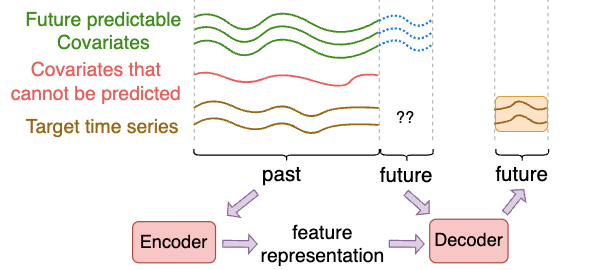}
\caption{Direct method using Seq2Seq models with encoder and decoder.} 
\label{encoder_decoder}
\end{center}
\end{figure}

Existing methods employing both past and future data for time series forecasting problems are mainly divided into two categories: (1) {\textbf{iterative methods} \cite{Rangapuram2018Deep,Salinas2020DeepAR} that iteratively predict one step at a time, and (2) {\textbf{direct methods}} \cite{Lim2021,Wen2017} that are trained to explicitly forecast the pre-defined horizons with sequence-to-sequence models (which originated from the speech translation domain \cite{Luong2015}). However, they have several limitations. The iterative methods consider the prediction output from the previous time step as the input for the next time step during the model training process. Such methods suffer from error accumulation caused by the multiplication of errors. 

In another direct strategy (see Fig.~\ref{encoder_decoder}), the Seq2Seq framework – encoder and decoder \cite{Qing2020} absorbs the historical data in the encoder and includes the predicted future covariates in the decoder. Such a strategy considers historical data and the predicted future covariates separately, probably causing the model to miss the past-future connections. Some researchers have added an attention layer \cite{Du2020,Yuan2020} in the Seq2Seq framework to capture more local or global information, but the prediction performance improves only slightly and fails to handle the inherent constraints of the Seq2Seq model.

In this work, we aim to address the existing limitations, and our five-fold contributions are listed below:
\begin{itemize}
    \item To avoid separately considering the past and future data, we propose a novel feature representation strategy called \emph{shifting}, which can contextually link the past with the predicted future covariate as an integrated input. \emph{Shifting} also makes it possible to use a single compact model to effectively combine both past and future data simultaneously.
    \item To improve the efficiency of the model, we introduce a \underline{para}llel framework composed of \underline{R}NN and \underline{CNN} (ParaRCNN) to capture complex time series dynamics. Note that ParaRCNN is a single and compact model compared to the Seq2Seq architecture.
    \item Our model can make multi-step predictions in a one-shot manner, which can avoid error accumulation in contrast to auto-regressive models.
    \item We adapt the \emph{skip connection} to facilitate improved learning since such a technique can maximize the usability of input features.
    \item We provide the model interpretability with the Grad-CAM algorithm to identify how each time step and feature contributes to the final predictions.
\end{itemize}

\section{Problem Formulation}
\label{problem_formulation}
Let ${\bf Z}^N_{1:t} = (z_1^n, z_2^n, ..., z_t^n)_{n=1}^{N} \in R^{t\times N}$ be $N$ univariate time series of target variables, where $z_t^n \in$ ${R}$ denotes the value of the $n$-th target variable at time $t$.
Let ${\bf X}^M_{1:t} = (x_1^m, x_2^m, \ldots, x_t^m)_{m=1}^{M} \in R^{t\times M}$ be $M$ the observed time-varying covariates that are measured until time $t$ and that cannot be predicted for the future.
Finally, let ${\bf Y}^Q_{1:t} = (y_1^q, y_2^q, \ldots, y_t^q)_{q=1}^{Q} \in R^{t\times Q}$ be the $Q$ time series for covariates measured until time $t$, but which can be reliably estimated for the near future; we let ${\bf Y}^Q_{t+1:t+k} \in R^{t\times Q}$ denote those predicted covariates $k$ time steps into the future.
We refer to these estimable variables as \emph{future predictable covariates}.
The goal of forecasting models is to compute the predicted trajectories of the target time series. We will refer to these as ${\bf \hat{Z}}^N_{t+1:t+k}$ ($k$ is the forecasting length) to distinguish it from the measured target time series. The computations assumes that the input data is heterogeneous and includes the historical data (target variables ${\bf Z}^N_{1:t}$, observed covariates ${\bf X}^M_{1:t}$, historical future predictable covariates ${\bf Y}^Q_{1:t}$), and predicted future covariates ${\bf Y}^Q_{t+1:t+k}$. Note that the term ``covariate'' in this paper refers to those exogenous time-varying covariates rather than time itself.

\section{Related Work}
\label{sec:relatedWork}
Traditional statistical methods learn the temporal patterns only based on historical data \cite{Chen2009ARIMA,Siami2018} of target variables themselves (see Eq. (\ref{Eq1})).
However, many approaches also aim to learn the dependent relationship between target variables and covariates, especially for the predicted future covariate \cite{Chen2020,Fan2019,Lim2021,Rangapuram2018Deep,Salinas2020DeepAR,Therneau2017,Wen2017}.
Related research can be mainly categorized into iterative methods using auto-regressive models and direct strategies that use sequence-to-sequence models.
We have:
\begin{equation}
\label{Eq1}
    {\bf \hat{Z}}^N_{t+i} = {\bf F}_{\theta}({\bf Z}^N_{1:t}, {\bf \hat{Z}}^N_{t+1:t+i-1}),
\end{equation}
where ${\bf F}_{\theta}(\cdot)$ is a prediction model with a set of learnable parameters $\theta$; ${\bf \hat{Z}}^N_{t+i}$ is the $N$ target variables $i$ time step into the future for $i={1,2,\ldots,k}$.

\textbf{Iterative methods.} The iterative strategy recursively uses a one-step-ahead forecasting model \cite{Dong2013,Yang2008} multiple times where the predicted value for the previous time step is used as the input to forecast the next time step.
A typical iterative framework is the DeepAR model \cite{Salinas2020DeepAR} from Amazon Research.
During the training process, to predict target values ${\bf \hat{Z}}^N_{t}$ at time step $t$, the inputs to the network are the covariates ${\bf Y}^Q_t$, the target values at the previous time step ${\bf Z}^N_{t-1}$, and the previous network output ${\bf h}_{t-1}$. 
Note that the previous target values are known during training.
During inference, measured target values ${\bf Z}^N_{t-1}$ are replaced by predicted target values ${\bf \hat{Z}}^N_{t-1}$ and then fed back to predict the next time step of ${\bf \hat{Z}}_{t+1}^N$ until the end of the prediction range.
A mathematical formulation of such forecasting methods is given in Eq. (\ref{Eq2}) using the notation in Section \ref{problem_formulation}.
Similar approaches were adopted in \cite{Lamb2016,Li2019Enhancing,Rangapuram2018Deep} using different backbones.
However, an inherent shortcoming of this method is that errors accumulate multiplicatively since later predictions depend on earlier predictions. 
\begin{equation}
\label{Eq2}
    {\bf \hat{Z}}^N_{t+i} = {\bf F}_{\theta}({\bf Z}^N_{1:t}, {\bf \hat{Z}}^N_{t+1:t+i-1}, {\bf X}^M_{1:t}, {\bf Y}^Q_{1:t+i}).
\end{equation}

\textbf{Direct methods.} The typical Seq2Seq framework for direct methods is shown in Fig \ref{encoder_decoder}. It deals with past and future data separately in the encoder and decoder components, respectively.
The encoder model learns the feature representation of past data, which is saved as context vectors in a hidden state.
The decoder model takes as input the encoder output and the additional future covariates to predict the future target values. 
Examples of this approach include the MQRNN model \cite{Wen2017} that used an LSTM as the encoder to generate context vectors, which are then combined with future covariates and fed into a multi-layer perceptron (MLP) to predict the future horizon.
Some efforts \cite{Du2020,Fan2019} have utilized a temporal attention mechanism between the encoder and the decoder. This architecture can learn the relevance of different parts of the feature representations from historical data by computing ``attentional'' weights. The weighted feature representations are then passed into the decoder to predict future time steps. 
Temporal Fusion Transformer \cite{Lim2021} combined gated residual networks (GRNs) and an attention mechanism \cite{Vaswani2017} as an additional decoder on top of the traditional encoder-decoder model. They used GRNs to filter unnecessary information and employed the additional decoder with an attention mechanism to capture long-term dependencies. Generally, the direct methods can be modeled as follows:
\begin{equation}
\label{Eq3}
\begin{aligned}
        {\bf H}_{t} = {\bf F}_{encoder}({\bf Z}^N_{1:t}, {\bf X}^M_{1:t}, {\bf Y}^Q_{1:t}),  \\
    {\bf \hat{Z}}^N_{t+i} = {\bf F}_{decoder}({\bf H}_{t}, {\bf Y}^Q_{t+1:t+i}).
\end{aligned}
\end{equation}
Direct methods that use the Seq2Seq framework with the encoder and decoder in series might be prone to miss some interactions between the past and future due to separate processing styles.
Moreover, the Seq2Seq framework is complicated and computationally time-consuming because of the use of two models -- the encoder and the decoder.
This provided the motivation for us to explore a compact model that simultaneously analyzes the measured past and the predictable future.

\section{Methodology}
In this section, we first illustrate the {\it shifting} strategy that fuses the past and future data in a structured way for an integrated feature representation.
Then we present the details of the proposed model architecture and discuss how it learns from the fused data and the skip connection technique. In this paper, we define a sliding window \cite{Gidea2018} (also called rolling window \cite{Li2014Rolling} or look-back window \cite{shi2022time}) of a certain length, $w$, as the input from the recent past, and to predict future time steps of length $k$.

\subsection{Data Fusion with Shifting}
\label{shifting}
To avoid dealing with the past and future data separately, we shift the covariates for the future period of interest (blue dotted trajectory in Fig. \ref{fig:data_fusion}) back by $s$ time steps, such that they are aligned and fused with all historical time series to produce distinct feature vectors. Then both the past and future data are fed into a single model together. Now the inputs are composed of all the past time series (target and covariates) aligned from time steps $t-w+1$ to $t$ with future predictable covariates from time steps $t-w+1+s$ to $t+s$.
Specifically, at each time step, we obtain a 4-tuple $(z_j,x_j,y_j, y_{j+s})$, which is input to a state cell in the RNN (Fig. \ref{fig:rnn}) or a filter kernel in the CNN (Fig. \ref{fig:cnn}), thus fusing the information from the historical data $(z_j,x_j,y_j)$ at time $j$ and future predictable covariates $y_{j+s}$ at time $j+s$.
The above design allows both the past and future to be considered in one single component of the model at the same time. The set of target variables, ${\bf Z}_{t+1:t+k}^N$ are predicted in the forecasting horizon from $t+1$ to $t+k$. 
The \emph{shifting} strategy is illustrated in Fig. \ref{fig:data_fusion} and modelled as Eq. (\ref{Eq4}) below:
\begin{align}
\label{Eq4}
    {\bf \hat{Z}}^N_{t+1:t+k} = {\bf G}_{\theta}({\bf Z}^N_{t-w+1:t}, {\bf X}^M_{t-w+1:t}, {\bf Y}^Q_{t-w+1:t}, {\bf Y}^Q_{t-w+1+s:t+s}),
\end{align}
where ${\bf G}_{\theta}(\cdot)$ is a function with learnable parameters ${\theta}$; and ${\bf Y}^Q_{t-w+1+s:t+s}$ is the future predictable covariates along with predictions from $s$ time steps into the future and then shifted back by $s$ time steps (green trajectories merged with dotted blue trajectories in Fig. ~\ref{fig:data_fusion}).
Note that the shifted future predictable covariates ${\bf Y}^Q_{t-w+1+s:t+s}$ and the single unified model given by ${\bf G}_{\theta}$ in Eq. (\ref{Eq4}) differentiate our method from the previous methods discussed in Section \ref{sec:relatedWork}. 
\begin{figure}[ht]
\begin{center}
\includegraphics[width=0.85\linewidth]{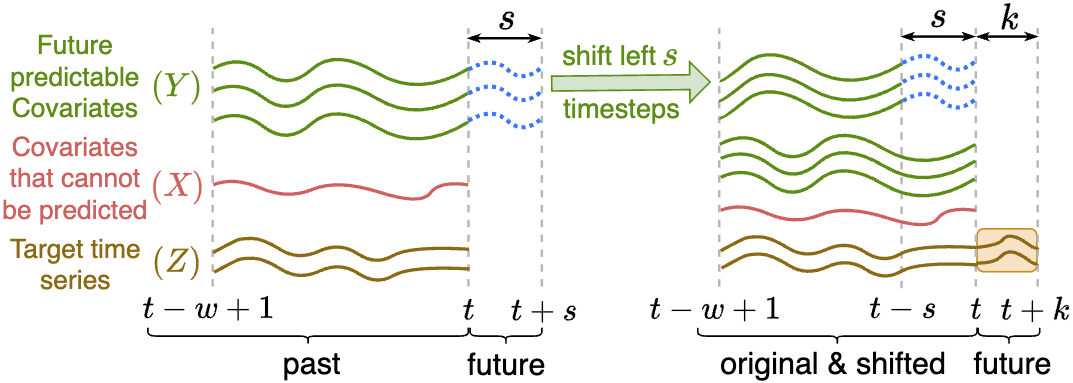}
\end{center}
\vspace{-3mm}
\caption{Input data transformed by the \emph{shifting} strategy. Left: Trajectories of all variables before transformation. Right: Original trajectories along with shifted future predictable covariates. Predicted output is the future $k$ time steps of the target variables.}
\vspace{-6mm}
\label{fig:data_fusion}
\end{figure}
\vspace{+3mm}
\subsection{Network Architectures}
With the input data transformed and fused (Fig. \ref{fig:data_fusion}, right) by the \emph{shifting} strategy, we develop a parallel framework composed of RNN and CNN, both of which are in a hierarchical structure.
As shown in Fig. \ref{fig:architecture}, both the number of filters for CNN and the number of units for RNN decrease over the layers to extract high-level time series dynamics. More specifically, since RNN and CNN learn the temporal dependency and dynamics in different mechanisms, we construct RNN and CNN in parallel, which benefits the model by capturing heterogeneous feature representations from input time series. 
Meanwhile, the \emph{skip connection} technique is utilized to enhance learning since it maximizes the usability of the input features.
Lastly, the fused input, the CNN output, and the RNN output are concatenated together and fed into a fully-connected layer to make the final predictions.
\begin{figure*}[ht]
\begin{center}
\includegraphics[width=0.95\linewidth]{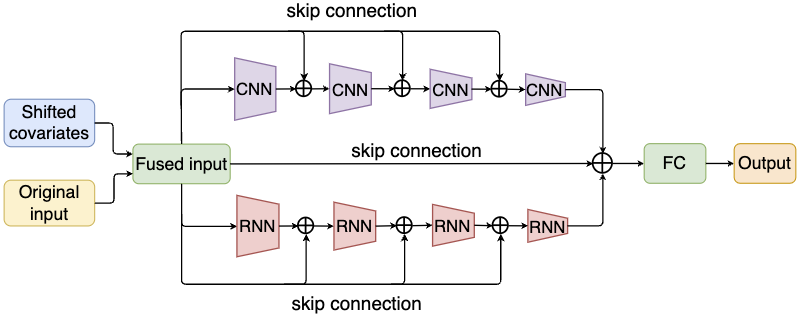}
\end{center}
\caption{Architecture of the proposed ParaRCNN model. There are 256, 128, 64, and 32 filters for CNN modules (Conv1D) and 128, 64, 32, and 16 units for RNN modules (SimpeRNN), respectively.}
\label{fig:architecture}
\end{figure*}
\vspace{-6mm}
\subsubsection{RNN with Shifting}
Recurrent Neural Networks (RNNs) learn the temporal dependency from input features in the recent past to future one or more target variables by recurrently training and updating the transitions of an internal (hidden) state from the last time step to the current time step. To predict the future $k$ time steps, the standard RNNs were further modified to remove the hidden states ${\bf h}_{t+1},{\bf h}_{t+2},..,{\bf h}_{t+k}$ to enable a one-shot prediction while avoiding the accumulation of prediction errors. As shown in Fig. \ref{fig:rnn}, we implement the RNNs with only $w$ hidden states in our paper. The predicted future covariates are shifted to the past by $s$ time steps and aligned with past data by the {\it shifting} strategy such that the input for each hidden state ${\bf h}_j$ at time $t=j$ is a 4-tuple (${\bf z}_j, {\bf x}_j, {\bf y}_j, {\bf y}_{j+s}$). Hierarchical RNNs queued in series (Fig. \ref{fig:architecture}) are expected to distill the high-level features from the input time series. At last, the RNNs generate the prediction for target variables $({\bf z}_{t+1}, {\bf z}_{t+2},\ldots, {\bf z}_{t+k})$ in a one-shot manner. The hidden states are recursively computed by:
\begin{equation}
\begin{aligned}
\label{Eq5}
    {h_j} & = {f}({\bf h}_{j-1}, {\bf z}_j, {\bf x}_j, {\bf y}_j, {\bf y}_{j+s}), \\
          & = tanh({\bf b} + {\bf U}^Th_{j-1} + {\bf W}^T({\bf z}_j, {\bf x}_j, {\bf y}_j, {\bf y}_{j+s})),
\end{aligned}
\end{equation}
where $f$ is an activation function (hyperbolic tangent function); ${\bf h}_j$ and ${\bf h}_{j-1}$ refer to the current and previous hidden states; ${\bf z}_j$, ${\bf x}_j$, and ${\bf y}_j$ represent the target time series, observed covariates, and predictable future covariates from the past $w$ time steps; ${\bf y}_{j+s}$ denotes the predicted future covariates from $k$ steps into the future; ${\bf U}, {\bf W}$ are weight matrices and ${\bf b}$ is the bias vector.

\begin{figure}[ht]
\begin{center}
\includegraphics[width=0.85\linewidth]{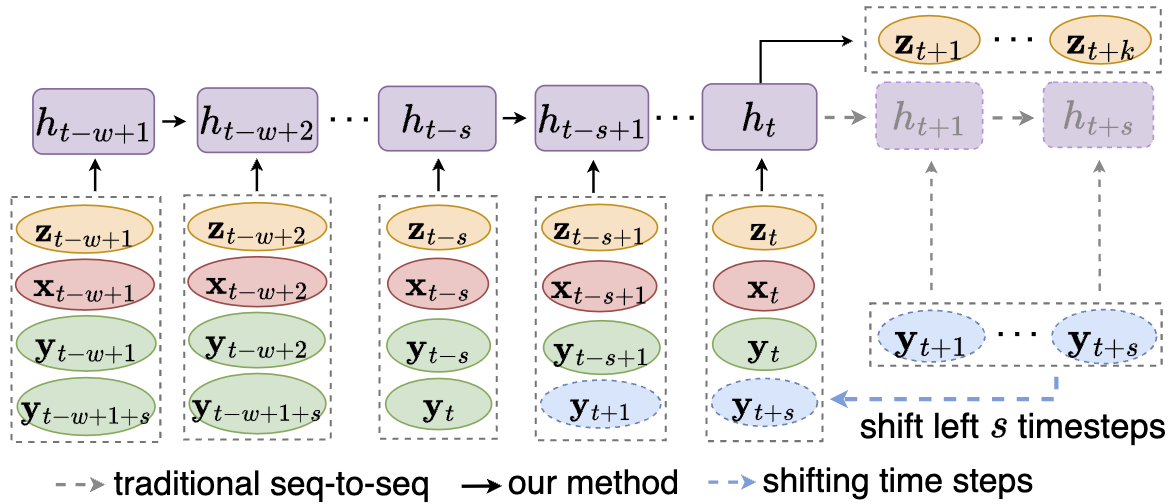}
\end{center}
\caption{The RNN architecture with the \emph{shifting} strategy. Dashed blue ovals represent predicted future covariates. Solid ovals are historical target variables and covariates. The last row has the shifted covariates.} 
\label{fig:rnn}
\end{figure}

\subsubsection{CNN with Shifting}
\begin{figure}[ht]
\begin{center}
\includegraphics[width=0.65\linewidth]{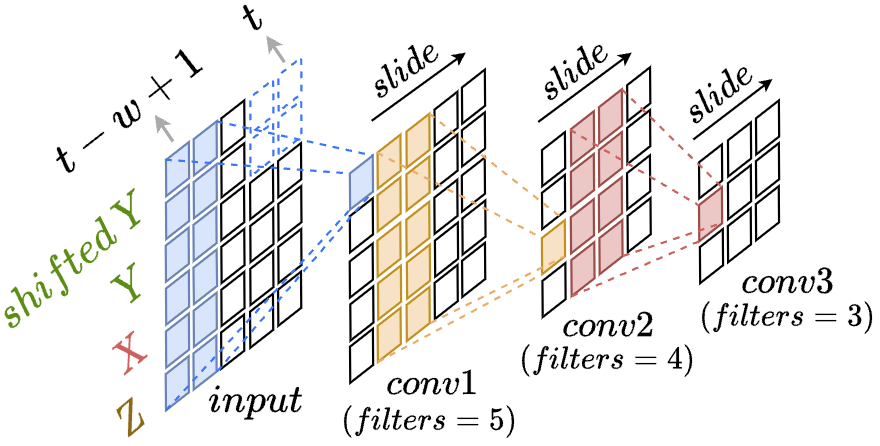}
\end{center}
\caption{CNN with 3 convolutional layers. Input includes all original variables and the shifted predicted future covariates. Each filter includes past and future information. Each row represents the convolution results with one filter.}
\label{fig:cnn}
\end{figure}
CNN is a popular model in the image processing field because of the powerful learning ability of {\it convolutional kernels} embedded inside. 2D-CNN is widely adopted to deal with images \cite{Yang2018} by moving 2-D convolutional kernels along the height and width dimension of each image. For multivariate time series, it consisted of multiple univariate time series that fundamentally are sequential in nature.  Therefore, 1-D convolutional kernels (also called filters) are used in our paper to learn the temporal and cross-feature dependency \cite{Tang2020}. We consider the multivariate time series as a matrix with the shape of (rows, columns) \cite{Zhang2019diagnosis} where the rows represent the time steps, and the columns represent the features that generally equal the number of time series dimensions. We also tried 2D-CNN, and the performance was not much different from 1D-CNN, but it needs more computation resources.

As Fig. \ref{fig:cnn} shows, the shifted future predictable covariates and the original observed data are simultaneously considered by the sliding 1-D convolutional kernels. In other words, each 1-D convolutional kernel learns from the historical data (past) and the predicted data $s$ time steps ahead (future). Such convolutional operations on both the history and predicted future input could be described by Eq. (\ref{Eq6}). Formally, a convolution operation between two convolutional layers is given by Eq. (\ref{Eq7}).  
\begin{equation}
\label{Eq6}
    {V}_{j} = {\sigma}(K_{j}{\circledast}({\bf Z}^N_{j:j+\Delta t}, {\bf X}^M_{j:j+\Delta t}, {\bf Y}^Q_{j:j+\Delta t}, {\bf Y}^Q_{j+s:j+\Delta t+s})),
\end{equation}
where $\circledast$ refers to the convolution operator; $K_j$ is a filter at the time $j$; $\Delta t$ is the length of segmented time series for the convolutioncomputations; $\sigma$ represents the activation function; $V_j$ is the output value at time $j$.
\begin{equation}
\label{Eq7}
    {a}_{j}^{l+1} = {\sigma}(b_{j}^{l} + \sum_{f=1}^{F^l}{K_{jf}^{l} \circledast a_{f}^{l}}),
\end{equation}
where $\circledast$ represents the convolution operator; $l$ indexes the layer, $f$ indexes the filter; $K_{j}$ is a filter at the time $j$; $F^l$ is the number of filters used in the $l^{th}$ layer; $\sigma$ denotes the activation function.

\subsubsection{Skip Connection}
It is used to train a deep neural network by copying and bypassing the input from the former layers to the deeper layers by matrix addition. ResNets add a skip-connection that bypasses the non-linear transformations with an identity function. For example, given a single image $x_0$ that is passed through subsequent convolutional layers, each layer implements a non-linear transformation $H(\cdot)$. The output of $l^{th}$ layer with skip connection looks as: 
\begin{equation}
    \label{Eq8}
    {x}_{l} = H_l(x_{l-1}) + x_{l-1}.
\end{equation}
DenseNets \cite{Huang2017} achieves skip connections by concatenation. In their work, for each layer, the feature maps of all preceding layers and their own feature maps are used as inputs into all subsequent layers by simple concatenation as shown in Eq. (\ref{Eq9}). There are $L(L+1)/2$ skip connections for the networks with $L$ layers.
\begin{equation}
    \label{Eq9}
    {x}_{l} = H_l(x_{0}, x_{1}, \dots, x_{l-1}),
\end{equation}%
where $x_{0}, x_{1}, \dots x_{l-1}$ denotes the concatenation of the feature maps produced in previous layers.
It shows how the $l^{th}$ layer considers the feature maps of all former layers as input. 
\begin{equation}
    \label{Eq10}
    {x}_{l} = H_l(x_{l-1}) + x_0.
\end{equation}
However, challenges persist with both strategies. ResNets hinder the skip connection because of the matrix addition, which needs the same dimension for both preceding and subsequent matrices.
DenseNets have a more complex structure with $L(L+1)/2$ connections as it conveys all former outputs to the latter layers.
U-Net models simply pass the original input once to the latter layers.
In our model, we adopt $L$ skip connections by bypassing the original input to every latter layer with concatenation (see Eq. (\ref{Eq10}) and Fig. \ref{fig:various_sc}d). Such a structure can facilitate the model by reusing the original input many times and learning it directly while avoiding the \emph{vanishing gradient} issue of deeper layers \cite{Li2018Visualizing}.

\begin{figure}[ht]
\begin{center}
\includegraphics[width=\linewidth]{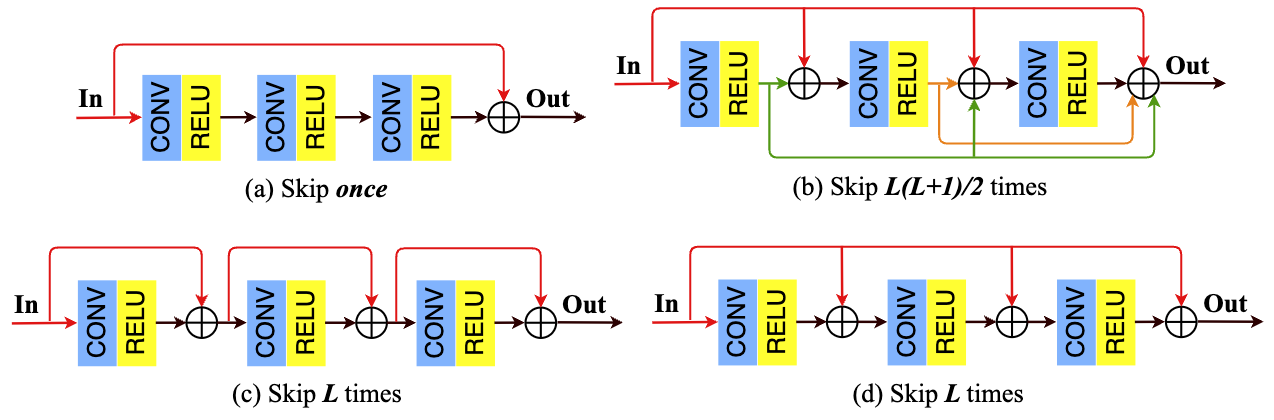}
\end{center}
\vspace{-6mm}
\caption{Various strategies for skip connection. We adopt the strategy of (d) in our paper and compare it with (a) Skip \emph{once} used in U-net \cite{Ronneberger2015}, (b) Skip $L(L+1)/2$ times used in DenseNet \cite{Huang2017}, and (c) the benchmark strategy skipping $L$ times.}
\label{fig:various_sc}
\end{figure}

\section{Experiments}
\subsection{Datasets}
Three real-world datasets were used for time series forecasting tasks. {\it Beijing PM2.5} and {\it Electricity price} datasets are publicly available from the Machine Learning Repository of the University of California, Irvine, and Kaggle repositories, respectively. The third one is the {\it Water Stage} dataset downloaded from the South Florida Water Management District (SFWMD) website.

{\bf Beijing PM2.5} It includes hourly observed data from January 1, 2010, to December 31, 2014. We consider PM2.5 as the target variable to predict, other variables such as dew, temperature, pressure, wind speed, wind direction, snow, and rain are covariates that can be predicted and can influence PM2.5 values. {\it PM2.5} $\in [0, 671]$ $\mu g/m^3$.

{\bf Electricity price} It has two hourly datasets from January 1, 2015, to December 31, 2018. {\tt Energy\_dataset.csv} includes energy demand, generation, and prices, while {\tt weather\_features.csv} gives the weather features temperature, humidity, etc. Electricity price is the target variable to predict, while the prior known predictable covariates are energy demand, generation, and weather features. {\it Electricity price} $\in [\$9.33, \$116.8]$ in this dataset.

{\bf Water stage} This is an hourly dataset from January 1, 2010, to December 31, 2020, and includes information on water levels, the height of gate opening, water flow values through the gate, water volumes pumped at gates, and rainfall measures. The water stage is the target variable while other variables are covariates. Rainfall, gate position, and pump control are future covariates that can be predicted. {\it Water stage} $\in [-1.25,4.05]$ feet in the dataset.

\subsection{Training \& Evaluation}
{\bf Our models}. We predict $k=24$ hours in the future with input windows of size $w=72$ hours and predicted future covariates in the same future horizons. We consider the entire target series ${\bf {Z}}^N_{t+1:t+k}$ as the ground-truth labels in supervised learning, which can allow one-shot forecasting to avoid the error accumulation of the traditional iterative prediction. For each dataset, we selected the first 80\% as the training set to train the models, and the remaining 20\% was chosen as the test set to evaluate the performance. {\it Max-Min normalization} shown as Eq. (\ref{normalization}) was used to squeeze the input data into $[0,1]$ to avoid possible data bias due to the different scales. We also used {\it early stopping} and {\it L1L2 regularization} to alleviate overfitting. There are several hyperparameters in our model. We set $\{16,32,64,128,256\}$ as the candidate numbers of internal units of RNNs and filters in CNNs. $\{\text{1e-3, 5e-4, 1e-4, 5e-5, 1e-5}\}$ was tested as the learning rate and regularization factor. The shifting length $s$ was validated with the range of $[1, w+k]$ (see Fig. \ref{fig:mae_rmse}). Open-source code can be accessed via the link \footnote{https://github.com/JimengShi/ParaRCNN-Time-Series-Forecasting.}.
\begin{equation}
\label{normalization}
    x' = \frac{x-x_{min}}{x_{max}-x_{min}}
\end{equation}
{\bf Baseline models}. DeepAR \cite{Salinas2020DeepAR} iteratively predicting future time steps was viewed as one of the baseline models. Seq2Seq approaches include MQRNN \cite{Wen2017} and Temporal Fusion Transformer (TFT) \cite{Lim2021}, which consider the past data and future covariates separately in the encoder-decoder framework. To validate the functionality of {\it shifting}, we also adapted baselines as a single branch in Fig. \ref{fig:architecture} (RNN or CNN) as backbones with the encoder-decoder framework (no {\it shifting}). We refer to them as RNN-RNN and CNN-CNN in Table \ref{tab:prediction_error}.

All models were trained by minimizing the loss function in Eq. (\ref{loss_function}), which describes the mean square error between predicted and ground-truth values. The training process is given as Algorithm \ref{training_algorithm}. The testing process is achieved by the trained model with the same data processing as the first 8 rows. Mean Absolute Errors (MAEs) and Root Mean Square Errors (RMSEs) are the metrics to evaluate the trained models. Each experiment was repeated 5 times with 5 random seeds. Table \ref{tab:prediction_error} reports the average results with an error bound.
\begin{equation}
\label{loss_function}
    L({\bf Z}, {\bf \hat{Z}}) = \frac{1}{\Phi} \sum_{\phi =1}^\Phi [({\bf Z}_{t+1,t+k}^N)^\phi-({\bf \hat Z}_{t+1,t+k}^N)^\phi]^2.
\vspace*{-6mm}
\end{equation}
\begin{algorithm}[tb]
\caption{Model Training}
\label{training_algorithm}
    \textbf{Input}: covariate time series: ${\bf X}_{1, T}, {\bf Y}_{1, T}$;\\
    \textcolor{white}{\textbf{Input}:} target time series: ${\bf Z}_{1, T}$, where $T$ is the total length of data set..\\
    \textbf{Parameter}: $w$: sliding window length, $k$: forecasting length, $s$: shifted length.\\
    \textbf{Output}: well-trained model
    \begin{algorithmic}[1] 
        \STATE {\it // construct training instance pairs}
        \STATE $D \gets \emptyset$ 
        \FOR{each available time point $w \le t \le T-s$}
        \STATE $S_{past} \gets \{{\bf X}_{t-w+1, t}, {\bf Y}_{t-w+1, t}, {\bf Z}_{t-w+1, t}\}$
        \STATE $S_{shifted} \gets \{{\bf Y}_{t-w+1+s, t+s}\}$
        \STATE $S_{target} \gets {\bf Z}_{t+1, t+k}$
        \STATE put a instance pair ($\{ S_{past}, S_{shifted}\}, S_{target}$) into $D$
        \ENDFOR
        \STATE {\it // train the model}
        \STATE initialize all learnable parameters $\theta$ for the model
        \REPEAT{}
        \STATE randomly selects a batch of instance pairs $D_b$ from $D$
        \STATE model outputs ${\bf \hat{Z}}_{t+1, t+k}$ for each batch
        \STATE finds $\theta$ by minimizing the loss function in Eq. (\ref{loss_function}) 
        \UNTIL{stopping criteria is satisfied}
        \STATE \textbf{return} well-trained model with the best parameters $\theta$
    \end{algorithmic}
\end{algorithm}
\vspace{-3mm}
\subsection{Hyperparameter Study}
\paragraph{\textbf{Shift length.}} Fig. \ref{fig:mae_rmse} shows the MAEs and RMSEs using ParaRCNN with different shifting lengths on the \texttt{Water-stage} dataset, which can help us to delineate the relationship between the shifting lengths and the model performance. We observed that $k \le s \le w$ generates better performance.\\
\vspace{-5mm}
\begin{figure}[ht!]
\begin{center}
\includegraphics[width=0.73\linewidth]{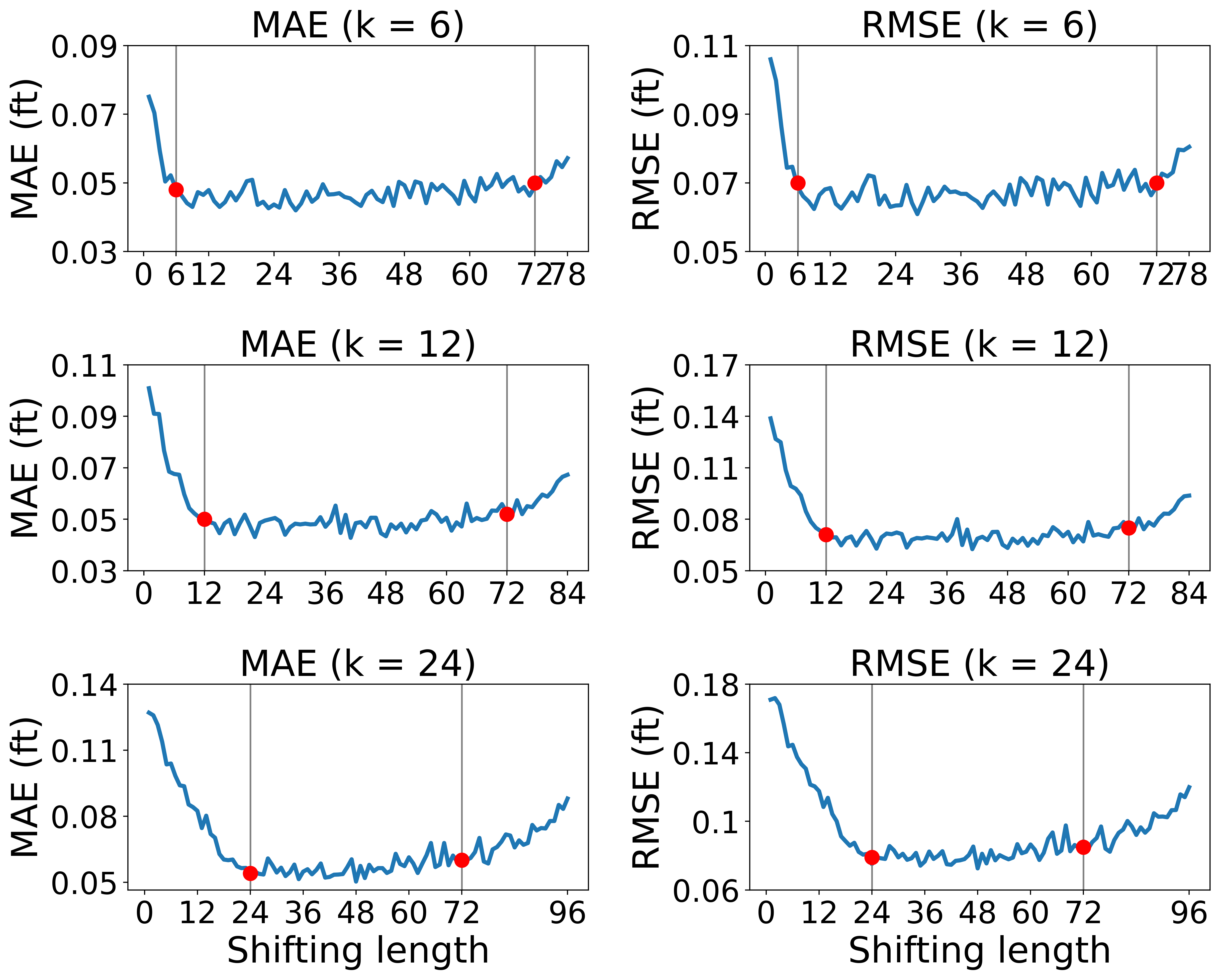}
\end{center}
\vspace{-4mm}
\caption{MAE \& RMSE for different forecasting lengths ($k$) and shift lengths ($s$). The left red point of each subplot represents the errors when $s=k$ while the right red point denotes the errors when $s=w$. ($w=72$ hours, $k=6, 12, 24$ hours.)}
\label{fig:mae_rmse}
\end{figure}

\paragraph{\textbf{Model layers.}} After ensuring the shift length $s=k$, we try to analyze the best number of layers for the ParaRCNN model. We found 3 or 4 layers (see Fig. \ref{fig:layer_error}) perform the best for the datasets in our paper (3 layers for \texttt{Electricity} dataset, 4 layers for \texttt{Water-stage} and \texttt{PM2.5} dataset).

\subsection{Prediction Results}
The first 5 rows in Table \ref{tab:prediction_error} show the performance of the baseline models. Our models are listed in the last 5 rows. We use a single RNN architecture of RNN-RNN and apply \emph{shifting} to it as RNN-Shift. To test the effectiveness of \emph{skip connection}, we add it to RNN-Shift and call it RNN-Shift-SC. A similar process is applied to CNN-Shift and CNN-Shift-SC. At last, we propose ParaRCNN (see Fig. \ref{fig:architecture}) by combining RNN and CNN in parallel with both \emph{shifting} and \emph{skip connection} techniques. Compared with baseline models in Table \ref{tab:prediction_error}, the performance of models with \emph{shifting} is comparable or slightly better than some baselines, while ParaRCNN achieves the best with the help of \emph{shifting} and \emph{skip connection}.
\vspace{-2mm}
\begin{table*}[ht!]
\centering
\caption{MAEs \&\ RMSEs with $k=24$ hours on the test sets.}
\small
\label{tab:prediction_error}
\begin{tabular}{ ll C{16mm}C{16mm} C{16mm}C{16mm} C{14.5mm}C{14.5mm} C{16.5mm}C{16.5mm}}
    \toprule
    &   \multirow{2}{*}{Methods}& \multicolumn{2}{c}{Beijing PM2.5} & \multicolumn{2}{c}{Electricity Price} & \multicolumn{2}{c}{Water Stage} \\
                         \cmidrule(lr){3-4}              \cmidrule(lr){5-6}               \cmidrule(lr){7-8}
           &                &   MAE &   RMSE         &   MAE &   RMSE                &   MAE &   RMSE            \\
                                                 \midrule
&  {\fontsize{8}{11}\selectfont MQRNN}             &  {\fontsize{8}{11}\selectfont 33.94$\pm$1.14} 
&  {\fontsize{8}{11}\selectfont 53.13$\pm$1.22}  &  {\fontsize{8}{11}\selectfont 3.48$\pm$0.14} 
&  {\fontsize{8}{11}\selectfont 4.69$\pm$0.19}   &  {\fontsize{8}{11}\selectfont 0.121$\pm$1e-2}  
&  {\fontsize{8}{11}\selectfont 0.156$\pm$4e-2}   \\

& {\fontsize{8}{11}\selectfont DeepAR}           &  {\fontsize{8}{11}\selectfont 36.57$\pm$0.72}
& {\fontsize{8}{11}\selectfont 57.75$\pm$0.98}   &  {\fontsize{8}{11}\selectfont 5.23$\pm$0.12}  
& {\fontsize{8}{11}\selectfont 6.59$\pm$0.18}    &  {\fontsize{8}{11}\selectfont 0.196$\pm$9e-3}  
& {\fontsize{8}{11}\selectfont 0.231$\pm$1e-2}   \\

& {\fontsize{8}{11}\selectfont TFT}              &  {\fontsize{8}{11}\selectfont 36.32$\pm$0.82}  
& {\fontsize{8}{11}\selectfont 60.13$\pm$1.37}   &  {\fontsize{8}{11}\selectfont 3.76$\pm$0.16} 
& {\fontsize{8}{11}\selectfont 5.52$\pm$0.24}    &  {\fontsize{8}{11}\selectfont 0.119$\pm$7e-3} 
& {\fontsize{8}{11}\selectfont 0.158$\pm$9e-3}   \\

& {\fontsize{8}{11}\selectfont RNN-RNN}          &  {\fontsize{8}{11}\selectfont 33.43$\pm$0.79}  
& {\fontsize{8}{11}\selectfont 52.43$\pm$1.15}   &  {\fontsize{8}{11}\selectfont 4.27$\pm$0.15} 
& {\fontsize{8}{11}\selectfont 5.72$\pm$0.26}    &  {\fontsize{8}{11}\selectfont 0.142$\pm$4e-3} 
& {\fontsize{8}{11}\selectfont 0.177$\pm$8e-3}   \\

& {\fontsize{8}{11}\selectfont CNN-CNN}          &  {\fontsize{8}{11}\selectfont 33.90$\pm$0.57}  
& {\fontsize{8}{11}\selectfont 53.15$\pm$1.22}   &  {\fontsize{8}{11}\selectfont 3.78$\pm$0.14} 
& {\fontsize{8}{11}\selectfont 5.08$\pm$0.21}    &  {\fontsize{8}{11}\selectfont 0.110$\pm$8e-3} 
& {\fontsize{8}{11}\selectfont 0.177$\pm$9e-3}   \\
                                                 \midrule    
& {\fontsize{8}{11}\selectfont RNN-Shift}        &  {\fontsize{8}{11}\selectfont 33.37$\pm$0.59}  
& {\fontsize{8}{11}\selectfont 52.96$\pm$1.27}   &  {\fontsize{8}{11}\selectfont 3.96$\pm$0.13} 
& {\fontsize{8}{11}\selectfont 5.23$\pm$0.24}    &  {\fontsize{8}{11}\selectfont 0.109$\pm$1e-2} 
& {\fontsize{8}{11}\selectfont 0.151$\pm$9e-3}   \\

& {\fontsize{8}{11}\selectfont RNN-Shift-SC}     &  {\fontsize{8}{11}\selectfont 31.90$\pm$0.55}  
& {\fontsize{8}{11}\selectfont 50.89$\pm$1.09}   &  {\fontsize{8}{11}\selectfont 3.49$\pm$0.12} 
& {\fontsize{8}{11}\selectfont 4.65$\pm$0.18}    &  {\fontsize{8}{11}\selectfont 0.071$\pm$7e-3} 
& {\fontsize{8}{11}\selectfont 0.096$\pm$7e-3}   \\

& {\fontsize{8}{11}\selectfont CNN-Shift}     &  {\fontsize{8}{11}\selectfont 33.55$\pm$0.46}  
& {\fontsize{8}{11}\selectfont 52.94$\pm$1.11}   &  {\fontsize{8}{11}\selectfont 3.85$\pm$0.14} 
& {\fontsize{8}{11}\selectfont 5.09$\pm$0.20}    &  {\fontsize{8}{11}\selectfont 0.131$\pm$8e-3} 
& {\fontsize{8}{11}\selectfont 0.158$\pm$9e-3}   \\

& {\fontsize{8}{11}\selectfont CNN-Shift-SC}     &  {\fontsize{8}{11}\selectfont 31.76$\pm$0.43}  
& {\fontsize{8}{11}\selectfont 50.61$\pm$1.08}   &  {\fontsize{8}{11}\selectfont 3.48$\pm$0.12} 
& {\fontsize{8}{11}\selectfont 4.69$\pm$0.17}    &  {\fontsize{8}{11}\selectfont 0.059$\pm$5e-3} 
& {\fontsize{8}{11}\selectfont 0.081$\pm$6e-3}   \\

& {\fontsize{8}{11}\selectfont \textcolor{blue}{\textbf{ParaRCNN}}} & {\fontsize{8}{11}\selectfont \textcolor{blue}{31.48$\pm$0.36}} 
& {\fontsize{8}{11}\selectfont \textcolor{blue}{49.97$\pm$0.89}}   &  {\fontsize{8}{11}\selectfont \textcolor{blue}{3.39$\pm$0.10}} 
& {\fontsize{8}{11}\selectfont \textcolor{blue}{4.60$\pm$0.13}}    &  {\fontsize{8}{11}\selectfont \textcolor{blue}{0.054$\pm$4e-3}} 
& {\fontsize{8}{11}\selectfont \textcolor{blue}{0.075$\pm$9e-3}}   \\

                                            \bottomrule
\end{tabular}
\end{table*}
\begin{figure}[ht!]
\begin{center}
\includegraphics[width=0.75\linewidth]{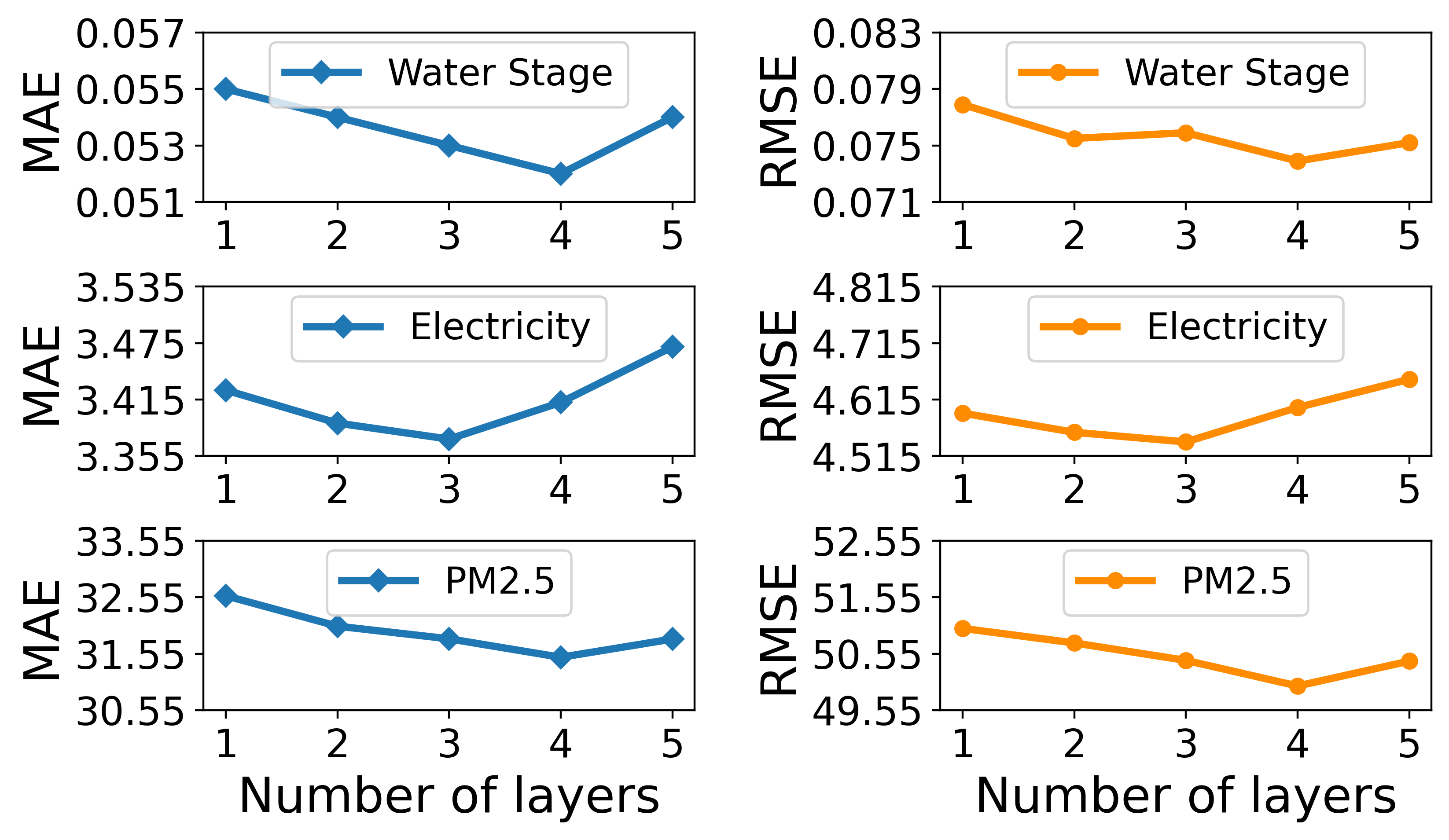}
\end{center}
\vspace{-5mm}
\caption{Model performance vs. Number of layers ($w=72$ hours, $k=24$ hours).} 
\label{fig:layer_error}
\end{figure}

\subsection{Skip Connection Study}
We apply \emph{skip connection} with different strategies (see Fig. ~\ref{fig:various_sc}) to the ParaRNN model. Taking an example of $L$ layers, there are five situations considered: (a) ${\bf One}$ skip in U-net \cite{Ronneberger2015}; (b) ${\bf L(L+1)/2}$ skips in DenseNet \cite{Huang2017}; (c) ${\bf L}$ skips; (d) ${\bf L}$ skips; and (e) {\bf No} skip connection. Fig. \ref{fig:various_sc_error} shows the performance of (e) without a skip connection is clearly much poor than others and (a-d) is roughly the same. The possible reason is that our network is a shallow one with only 4 layers. ${\bf L}$ and ${\bf L(L+1)/2}$ skips do not exist a big difference. However, the number of skip connections is indeed reduced from ${\bf L(L+1)/2}$ to ${\bf L}$.

\subsection{Model Explainability}
After the model was trained, we analyzed how much each time step and feature contribute to the final outputs. With the Grad-CAM algorithm \cite{Selvaraju2017}, we first compute the gradient of the target values with respect to the feature map activations of the concatenated layer. These gradients flow back over the input of shape (time steps $\times$ features) to obtain the neuron importance weights (see Fig. \ref{fig:model_interpretability} in Appendix A). The water stage at S1, S25A, S25B, and S26 are target values to predict. The first 19 rows are original past input and the last 9 rows shifted covariates (shifted future covariates are from 48 to 72).  It shows our model pays more attention to these future covariates since target variables in the future horizon have a dependent relationship with them.
This is as described in Section \ref{shifting}. The water stages at different stations are more correlated because these stations are adjacent to the ocean and water stages are changing with the trend of the tide (WS\_S4). We visualize each time series in Appendix B (see Figs. \ref{fig:vis_original_10_19} and \ref{fig:vis_shifted}) providing better observations for readers.
\vspace{-3mm}
\begin{figure}[ht!]
\begin{center}
\includegraphics[width=0.7\linewidth]{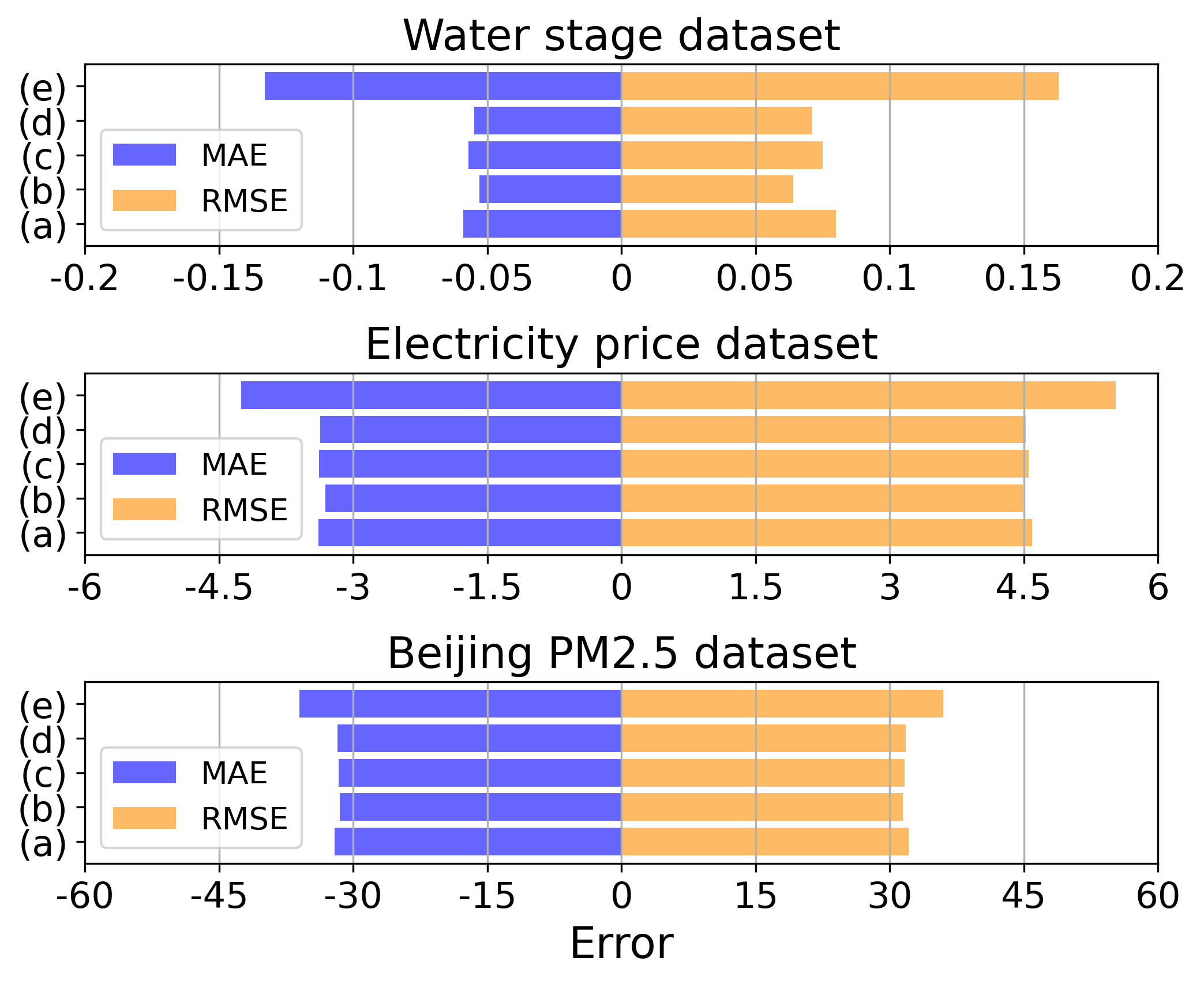}
\end{center}
\vspace{-5mm}
\caption{MAE \& RMSE for various strategies to implement {\it skip connection}. (a) ${\bf One}$ skip in U-net \cite{Ronneberger2015}; (b) ${\bf L(L+1)/2}$ skips in DenseNet \cite{Huang2017}; (c) ${\bf L}$ skips as one benchmark; (d) ${\bf L}$ skips we are using in our paper; and (e) {\bf No} skip connection. } 
\label{fig:various_sc_error}
\end{figure}
\vspace{-7mm}
\section{Discussion and Conclusions}
We have demonstrated with experiments that the utilization of future covariates can enhance performance. The model explainability shows their importance from another point of view. To take the advantage of future covariates, the proposed data fusion method, \emph{shifting}, can generate comparable or slightly better performance with a single compact model. Besides, our experiments delineate an appropriate range of the \emph{shift} length (see Fig. \ref{fig:mae_rmse}). When $s<k$ or $s>w$, considerably lower performances occur since the models only get to utilize some of the predicted covariates from the future $k$ time steps. However, when $s>k$, either the performance is flat or deteriorates as $s$ is increased. We observe that $k \le s \le w$ generates better performance since all predicted future covariates in the forecasting horizon are included. The variations for $k \le s \le w$ are too small to be significant.

\emph{Skip connection} can further improve the model performance.
Our implementation strategy that presents the original input to each subsequent layer generated roughly the same or better performance when compared with other strategies.
Our \emph{ParaRCNN} model equipped with \emph{shifting} and \emph{skip connection} techniques consistently outperformed all other models in our paper. 


%
%
%
\bibliographystyle{splncs04}
\bibliography{mybibliography}

\newpage

\section{Appendix}
\subsection{Model Explainability}
\label{model_explainability}
We provide the explainability of the trained model using the Water Stage dataset. The following figure shows how important each feature and each time step is for the final predictions.
\begin{figure}[H]
\begin{center}
\includegraphics[width=\linewidth]{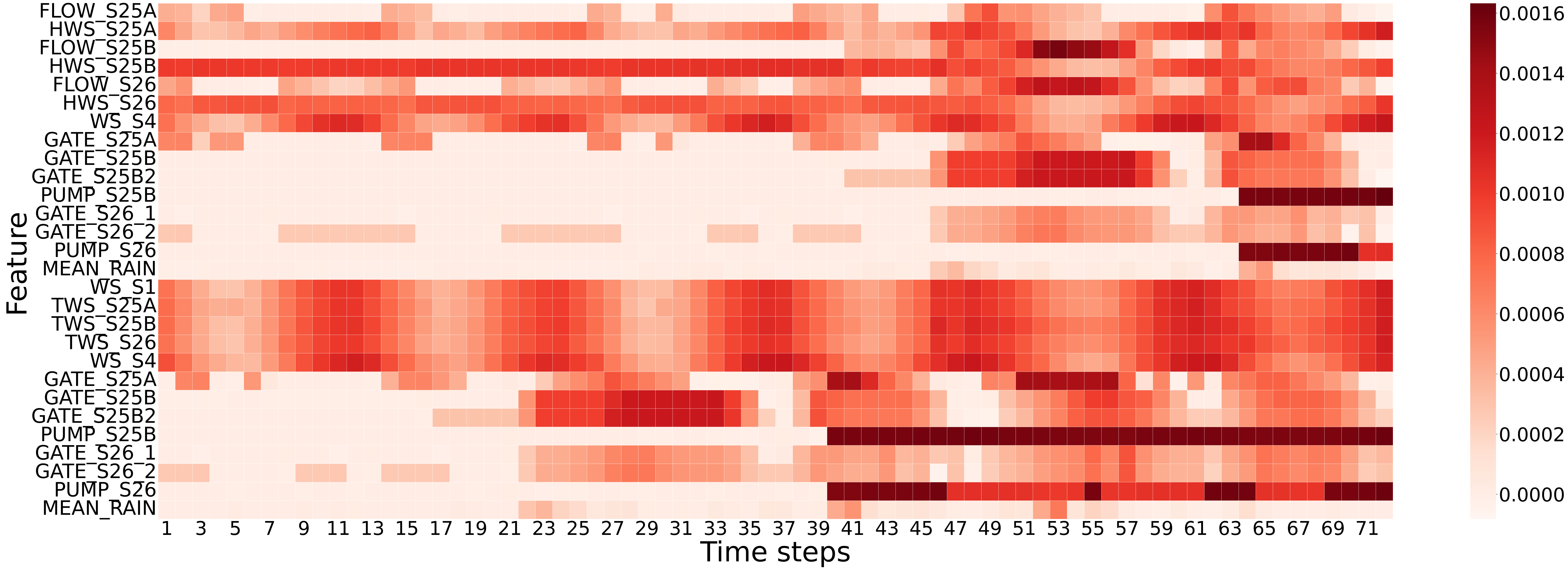}
\end{center}
\caption{Importance weights (feature vs. time step) with Grad-CAM algorithm.} 
\label{fig:model_interpretability}
\end{figure}

\newpage
\subsection{Visualization of Time Series}
\label{visualization_ts}
We visualize the time series in Fig. \ref{fig:model_interpretability} below. The unit of each feature is ignored. We refer readers to see Fig. \ref{fig:data_fusion} for better understanding.
\begin{figure}[ht!]
\begin{center}
\includegraphics[width=0.65\linewidth]{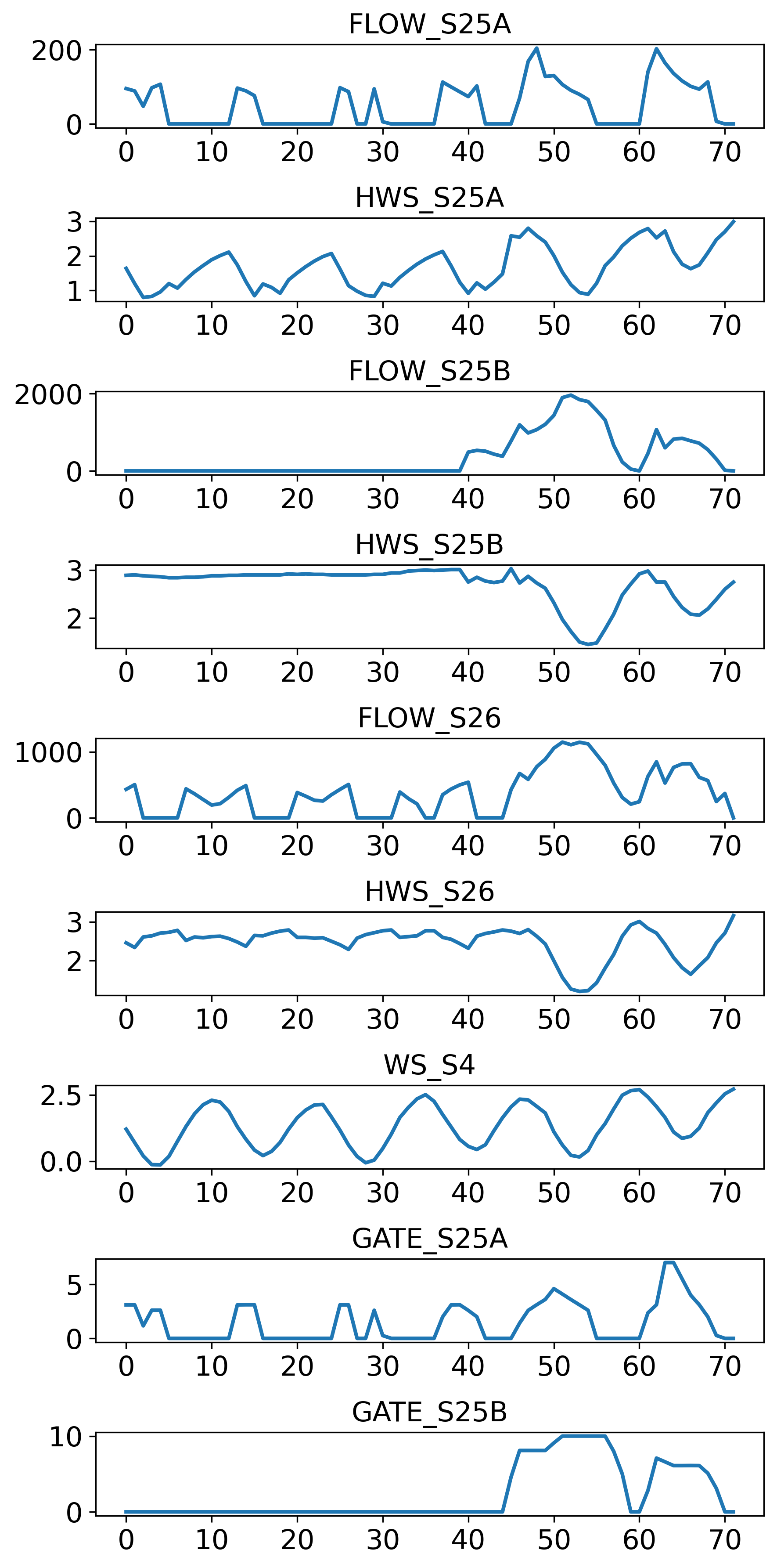}
\end{center}
\label{fig:vis_original_1_9}
\end{figure}

\begin{figure}[ht!]
\begin{center}
\includegraphics[width=0.65\linewidth]{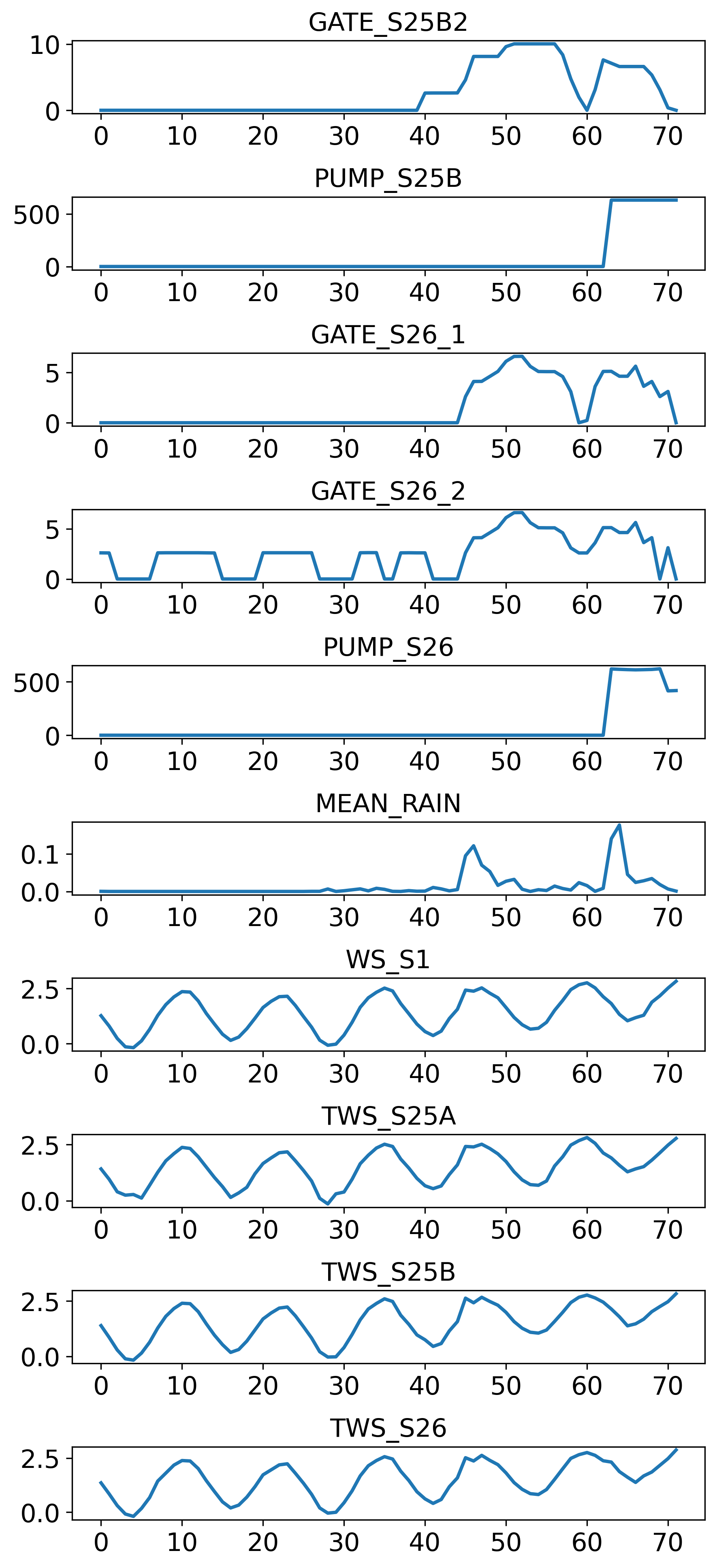}
\end{center}
\caption{Visualization of target variables and covariates from the past.} 
\label{fig:vis_original_10_19}
\end{figure}

\begin{figure}[ht!]
\begin{center}
\includegraphics[width=0.65\linewidth]{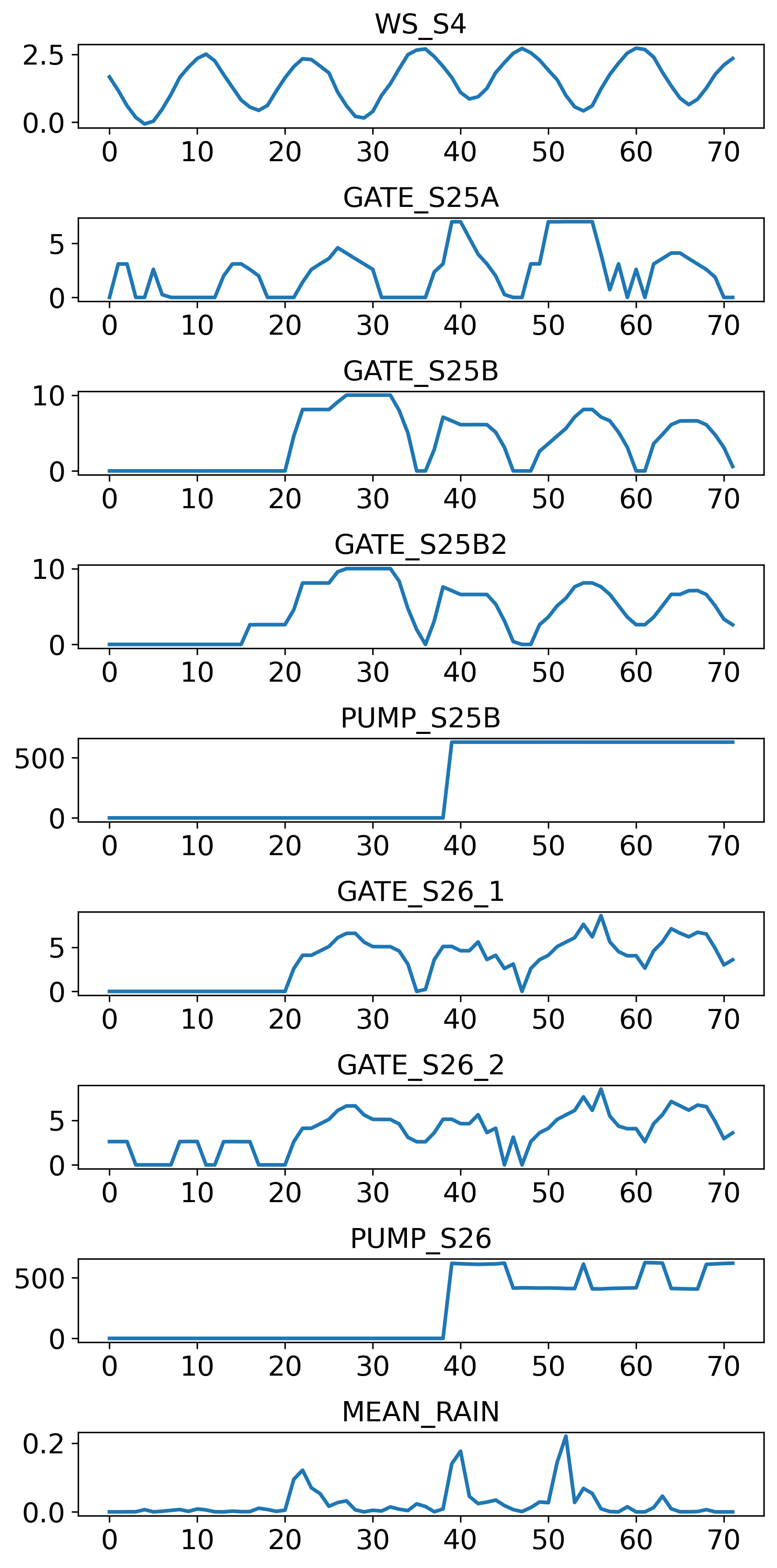}
\end{center}
\caption{Visualization of shifted future predictable covariates.} 
\label{fig:vis_shifted}
\end{figure}

\end{document}